\pdfoutput=1

\documentclass[11pt]{article}

\usepackage[]{acl}

\usepackage{times}
\usepackage{latexsym}

\usepackage{graphicx}
\usepackage{amsmath}
\usepackage{amsthm}
\usepackage{amssymb}
\usepackage{multirow}
\usepackage{subfigure}
\usepackage{algorithm}
\usepackage{algorithmic}

\usepackage[T1]{fontenc}

\usepackage[utf8]{inputenc}

\usepackage{microtype}

\title{Continual Machine Reading Comprehension via \\ Uncertainty-aware Fixed Memory and Adversarial Domain Adaptation}

\author{Zhijing Wu$^{1,2}$ , Hua Xu$^{1,2} \thanks{$\;$ Hua Xu is the corresponding author.}$ , Jingliang Fang$^{1,3}$ \and Kai Gao$^{3}$\\
        $^1$State Key Laboratory of Intelligent Technology and Systems, \\
        Department of Computer Science and Technology, Tsinghua University, Beijing 100084, China \\
        $^2$Beijing National Research Center for Information Science and Technology(BNRist) \\
        $^3$School of Information Science and Engineering, Hebei University of Science and Technology \\
  \texttt{wuzj18@mails.tsinghua.edu.cn} \\
  \texttt{xuhua@tsinghua.edu.cn}
        }

\begin{document}
\maketitle
\begin{abstract}
Continual Machine Reading Comprehension aims to incrementally learn from a continuous data stream across time without access the previous seen data, which is crucial for the development of real-world MRC systems.
However, it is a great challenge to learn a new domain incrementally without catastrophically forgetting previous knowledge.
In this paper, MA-MRC, a continual MRC model with uncertainty-aware fixed \textbf{M}emory and \textbf{A}dversarial domain adaptation, is proposed.
In MA-MRC, a fixed size memory stores a small number of samples in previous domain data along with an uncertainty-aware updating strategy when new domain data arrives.
For incremental learning, MA-MRC not only keeps a stable understanding by learning both memory and new domain data, but also makes full use of the domain adaptation relationship between them by adversarial learning strategy.
The experimental results show that MA-MRC is superior to strong baselines and has a substantial incremental learning ability without catastrophically forgetting under two different continual MRC settings.



\end{abstract}

\section{Introduction}
Recently, Machine Reading Comprehension (MRC) has attracted wide attention and achieved remarkable success when solving specific tasks in stationary environments, such as answering factual questions with wikipedia articles or answering narrative questions with web search logs \cite{SeoKFH17,DBLP:conf/emnlp/SeonwooKHO20,DBLP:conf/aaai/0001YZ21,DBLP:journals/kbs/WuX20}.
However, the answering scenario changes over time in real-world applications.
For example, the dialog system should continuously adapt to new user requirements \cite{DBLP:conf/www/AbujabalRYW18,DBLP:conf/emnlp/MadottoLZMCLYCF21}.
In this paper, we focus on one of the most typical scenario changes for MRC tasks: the domain data shift.
Existing stationary-trained MRC systems are usually trained with in-domain data but are applied to new domain data \cite{DBLP:conf/acl-mrqa/FischTJSCC19}.
Therefore, it is necessary to build a non-stationary MRC that continually learns with incremental domain data.

\begin{figure}[t]
\centering
\includegraphics[width=0.95 \linewidth]{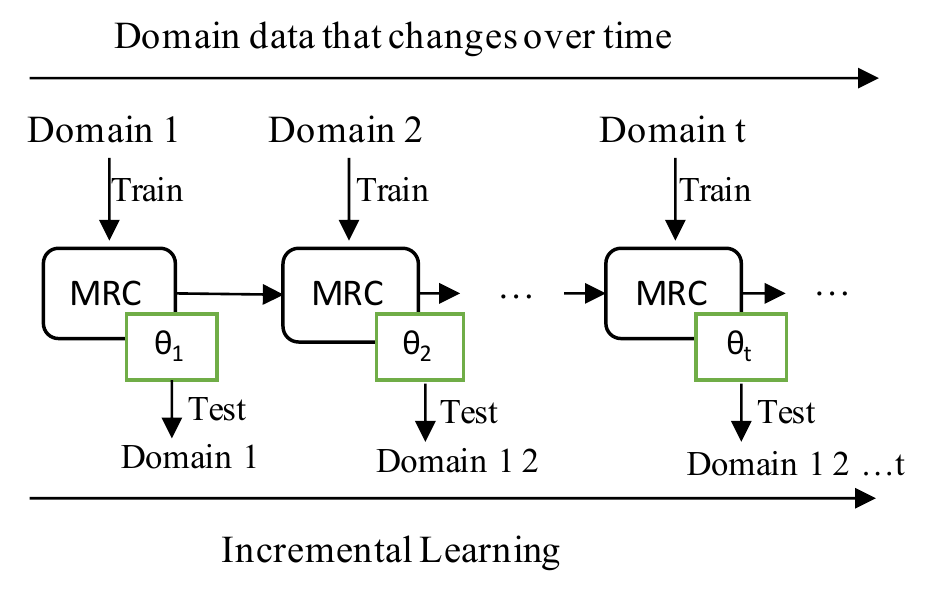}
\caption{Illustration of a continual MRC task.}
\label{example}
\end{figure}

We formulate such a challenging problem as \textit{Continual MRC} task, which is required to incrementally learn over sequential domain data and perform well on all seen domain data.
Figure \ref{example} illustrates the incremental learning and testing processing.
In this scenario, the MRC system can only trained on the latest coming domain data without access the previous seen data.
To tackle this issue, if we directly fine-tune the MRC model on each new incoming domain, the performance on earlier domains will significantly drop \cite{DBLP:conf/cikm/SuGZFLC20}. 
Another naive approach is to retrain the whole MRC model from scratch, but it is costly and time-consuming.
Hence it is a great challenge for incrementally learning without largely forgetting previously acquired knowledge.

Existing studies for continual MRC can mainly be divided into three categories.
The first class is model expansion techniques that design domain-individual classifier for each in coming domain \cite{DBLP:conf/cikm/SuGZFLC20}.
However, it is expensive and unpractical in real-world.
The second class borrows the regularization idea, which utilizes an additional loss term to aid knowledge consolidation when learning new domains.
For example, \citet{DBLP:conf/cikm/SuGZFLC20} added a penalty that restricts the change of important parameters to prevent forgetting previous knowledge.
The third class is episodic memory based methods.
For example, \citet{DBLP:conf/nips/dAutumeRKY19} introduced a key-value memory module that stored previously seen examples for sparse experience replay and gradient-based local adaptation.
\citet{DBLP:conf/www/AbujabalRYW18} proposed template-based Never-Ending KB-QA that learned new templates by capturing new syntactic structures with a semantic similarity function between questions and user feedback.

However, the above methods still have limitations, mainly including two aspects.
On the one hand, to prevent catastrophic forgetting, these methods only design consistent constraints of model output or gradient for previous and new domains, while ignoring the domain adaptation relationship between them.
However, transfer learning can help the MRC model generalize to other domains. 
On the other hand, the memory update strategy for continual MRC is limited.
Some previous work stores fixed examples for each incoming domain.
It greatly grows the number of samples kept in memory and leads to expensive costs.
Other methods that limit the maximum number of memory for all seen domains usually update the memory by random sampling, ignoring the forgotten degree of different samples.
In fact, the continual model should pay more attention to samples that are more likely to be forgotten.

To handle the above limitations, this paper proposes MA-MRC, an incremental model that solves continual MRC task via Uncertainty-aware fixed \textbf{M}emory and \textbf{A}dversarial Domain Adaptation.
Concretely, MA-MRC
1) introduces a fixed-size memory to store a small number of samples in previous domain, which are later periodically replayed when learning new domain;
and 2) updates the memory with an uncertainty-aware strategy that takes the forgotten degree of previous data into account;
3) leverages an adversarial learning strategy to make full use of the domain adaptation relationship between different domains with a domain discriminator, so as to help generalize and avoid overfitting very small memorized examples.
The intuition behind this is to mimic the human learning process that replays the memory and adapts to new domains.

The key contributions of this work are:
(1) This paper proposes a continual MRC model, MA-MRC, which learns new domain data incrementally.
(2) Applying uncertainty-aware \textbf{M}emory and \textbf{A}dversarial learning and to MRC model contributes to strong incremental learning ability.
(3) The experimental results on two different continual MRC settings indicate that MA-MRC obtaines good incremental learning ability without largely forgetting and significantly outperforms strong baselines.

\section{Related Work}
\subsection{Continual Learning}
Continual Learning (CL) mainly aims to overcome the catastrophic forgetting problem when learning on sequential new tasks incrementally \cite{FRENCH1999128}.
Existing work follows three directions: architectural, regularization, and memory-based approaches.
The architectural methods change the network's architecture and add task-specific parameters, e.g., Dynamically Expandable Network \cite{DBLP:conf/iclr/YoonYLH18} and Reinforced Continual Learning \cite{DBLP:conf/nips/XuZ18}.
The regularization-based techniques aid knowledge consolidation when learning new tasks.
For instance, EWC and Online EWC \cite{kirkpatrick2017overcoming,DBLP:conf/icml/Schwarz0LGTPH18} slow down the learning of parameters important for previous tasks.
The third class is to save previous samples and learn a new task with a forgetting loss defined on them, 
such as GEM \cite{DBLP:conf/nips/Lopez-PazR17}, A-GEM \cite{DBLP:conf/iclr/ChaudhryRRE19}, DER \cite{DBLP:conf/nips/BuzzegaBPAC20}, and MER \cite{DBLP:conf/iclr/RiemerCALRTT19}.

\textbf{CL in MRC.}
Few previous studies apply continual learning to MRC.
\citet{DBLP:conf/cikm/SuGZFLC20} adapted EWC method and enlarged the MRC architecture when a new domain arrives.
\citet{DBLP:conf/cikm/SuGZFLC20} added a penalty regularization that restricts the change of important parameters to prevent forgetting.
\citet{DBLP:conf/nips/dAutumeRKY19} and \cite{DBLP:conf/www/AbujabalRYW18} designed episodic memory based methods that store training samples from previously seen data, which are later rehearsed to learn new domains.
In this paper, we solve the continual MRC problem of incrementally learning over sequential domains, and build our continual model based on the above memory-based and penalty regularization paradigm.


\begin{figure*}[t]
  \subfigure[Initial Training.]{
  \includegraphics[width=0.25 \linewidth, height=0.56 \linewidth]{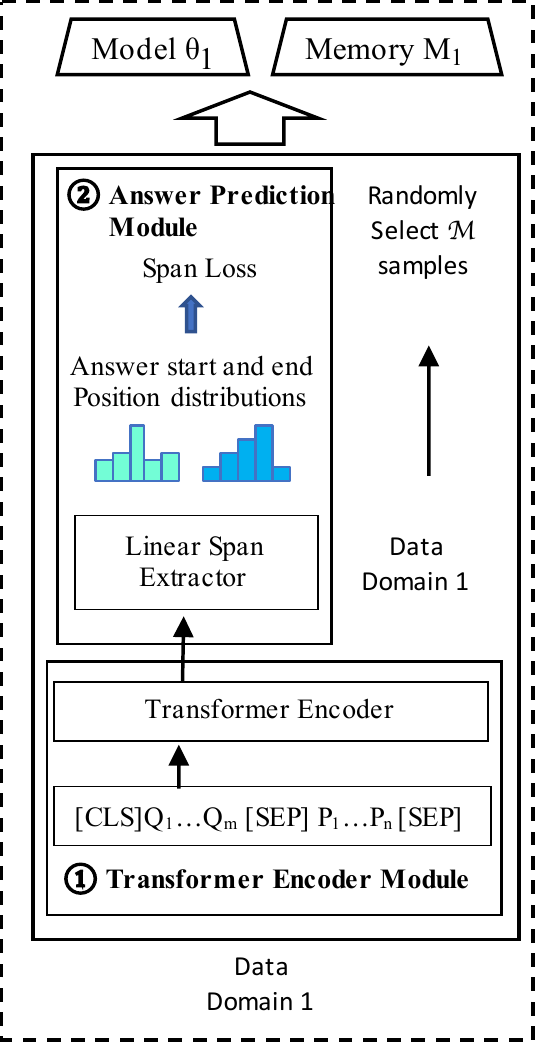}
  \label{frameworka}
  }
  \subfigure[Incrementally Learning for New Domain Data.]{
  \includegraphics[width=0.74 \linewidth, height=0.56 \linewidth]{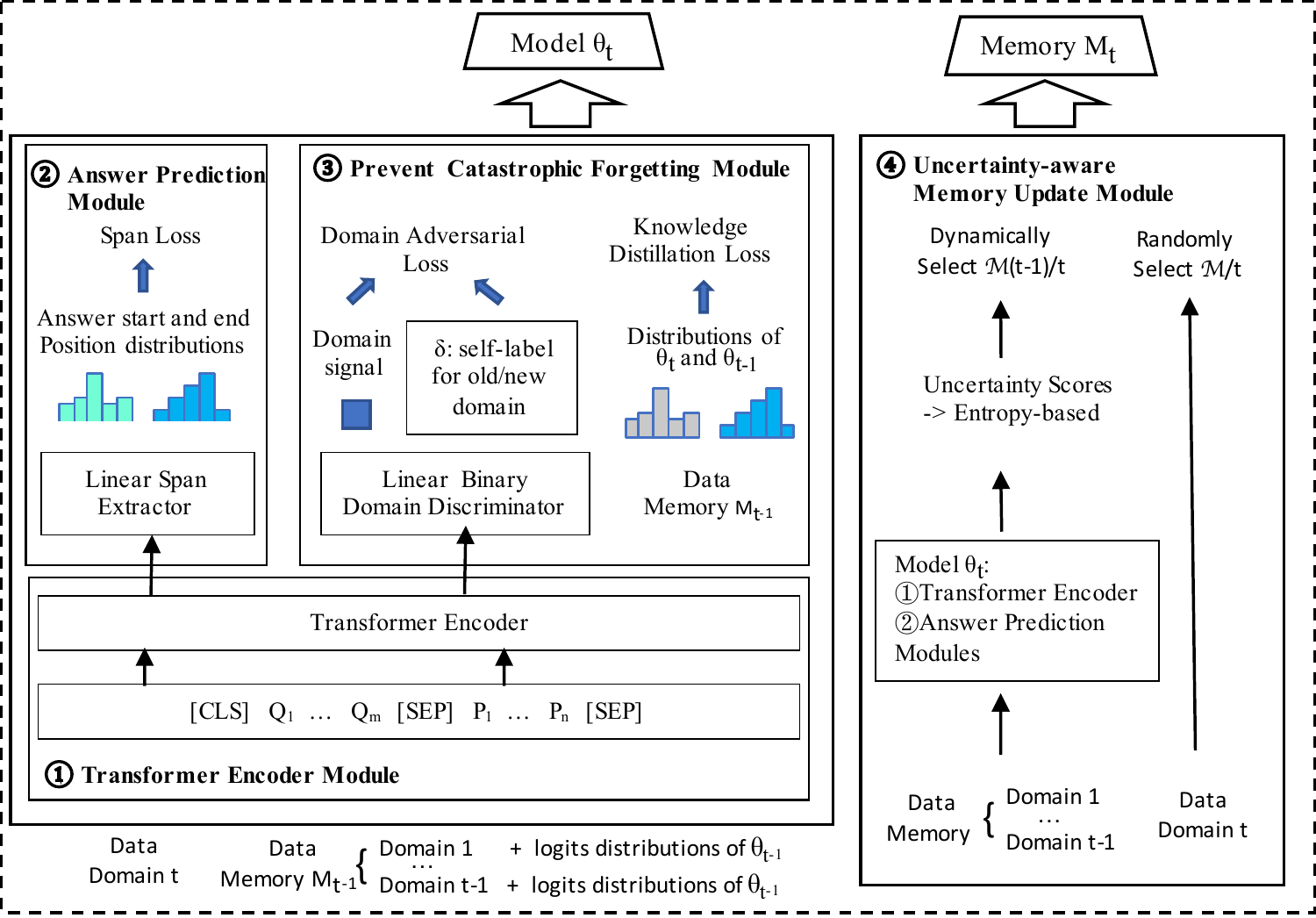}
  \label{frameworkb}
  }
  \caption{The overview of the proposed MA-MRC framework with sequential domain data.}
  \label{framework}
\end{figure*}

\subsection{Domain Adaptation}
Domain adaptation aims to learn discriminative feature features and generalize to other new domains and is usually achieved by learning domain-invariant features \cite{DBLP:journals/ml/Ben-DavidBCKPV10,DBLP:journals/jmlr/GaninUAGLLML16}.
For MRC under domain shift, previous methods \cite{DBLP:conf/emnlp/WangGLLGW19, DBLP:conf/aaai/CaoFYZ20,DBLP:conf/acl-mrqa/LeeKP19} reduce domain discrepancy by a discriminator network that is trained to distinguish features of the target from source domains.
However, recent work usually transfers the model from source to target.
In this paper, we explore the domain adaptation under continual setting.

\section{Proposed Method}
\subsection{Problem Statement}
For the task of continual MRC, we assume that the learning framework has access to streams of MRC data from $T$ different domains, denoted by $Data=\{D_1, D_2, ..., D_T\}$.
Each domain data, e.g. $D_t=\{P_n,Q_n,A_n\}_{n=1}^{|D_t|}$, consists of a series of $<$passage, question, answer$>$ triples, where $|D_t|$ is the sample number of $D_t$.
The MRC model is required to continually learn over each incoming domain data.
More specifically, at each step $t$, the MRC model only observes $D_t$ domain data, and is required to perform well on both the $t$-th domain data and previous $t-1$ domain data.
Hence, after training on $D_t$, the model will be evaluated on all seen $t$ domains.
To make the MRC model perform well on previous tasks, during the continual learning process, a memory $M$ is to set to store a small number of samples in previous domain data in a fixed-size memory $\mathcal{M}\ll|D_{old}|$.

This paper focuses on the span extractive MRC. 
Inspired by \citet{DBLP:conf/cikm/SuGZFLC20}, we perform two different continual domain settings: CDA-C and CDA-Q, which define different domains according to paragraph type and question type, respectively.

\subsection{Method overview}
Figure \ref{framework} shows the overview of MA-MRC.
As shown in Figure \ref{frameworka}, we first train a backbone MRC model (Transformer Encoder and Answer Prediction modules) on the initial domain data and randomly select $\mathcal{M}$ training samples as the initial memory $M$.
Figure \ref{frameworkb} shows the process of learning new domain data.
When the $t$-th domain data is arising, MA-MRC synchronously
1) fine-tunes the backbone MRC model with both the $t$-th domain and memory data;
2) adversarially learns domain-invariant and transfer representations via a domain discriminator that distinguishes memory from current data, so as to generalize well on new domain and avoid overfitting very small memory samples;
3) utilizes knowledge distillation to encourage the model to mimic its original responses for past domain samples.
After finishing training at $t$-th step, we update the memory with an uncertainty-aware sampling strategy that focuses on remembering what the model most needs.
More details about the training process are shown in Algorithm \ref{algorithm}.

\begin{algorithm}[t] 
\caption{Incremental Training for MA-MRC} 
\begin{algorithmic}[1] 
  \REQUIRE ~~\\ 
    T Training domain data $Data=\{D_1, D_2, ..., D_T\}$ \\
    A memory ${M}$ of size $\mathcal{M}$ \\
  \ENSURE ~~\\ 
    The Continual MRC model $\theta_T$ \\
    // \textit{Initial Training}
    \STATE Train the backbone MRC model $\theta_{1}$ with $D_1$ \\
    \STATE $M \gets Random Sampling (D_1, \mathcal{M})$ \\
    // \textit{Incrementally Learning for New Domain}
  \FOR{$t=2,...,T$}
    \STATE \textbf{Define} the domain discriminator $\theta_{D}$
    \STATE \textbf{Incrementally learning} $\theta_t \gets$ update $\theta_{t-1}$ and $\theta_{D}$ with $\bigtriangledown\mathcal{L}_\text{final}$ on $D_t,{M}_{t-1}$
    \STATE Calculate need importance based on uncertainty for each sample in ${M}_{t-1}$
    \STATE ${M}_t \gets Weight Sampling ({M}_{t-1}, \frac{\mathcal{M}(t-1)}{t}) + Random Sampling (D_t, \frac{\mathcal{M}}{t})$ \\
  \ENDFOR
  \RETURN {the final MA-MRC model $\theta_{T}$};
\end{algorithmic}
\label{algorithm}
\end{algorithm}

\subsection{Initial Training}
\label{backbone}
Note that there is only one domain data at the beginning, so the initial training with $D_1$ can be seen as a special case of formal extractive MRC tasks.
Therefore, we build a normal backbone MRC model $\theta_1$ (a standard BERT-MRC model with Transformer Encoder and Answer Prediction Modules) on the first domain data $D_1$.
We initialize a fixed-size memory $M_1$ to keep previous training samples that are periodically replayed while learning the new domain.

\subsubsection{Prepare Backbone model}
\noindent
\textbf{Transformer Encoder Module:}
First, a pre-trained $L$ Transformer encoder blocks is used to convert the input sequence $S=[\langle{CLS}\rangle,Q,\langle{SEQ}\rangle,P,\langle{SEQ}\rangle]$ into contextual representations.
Then, the last block output $H_L=\text{BERT}(S) \in\mathbb{R}^{(l)\times h}$ is taken as the contextual representation, where $h$ is the hidden dimension of BERT, $l$ is the sequence length.

\noindent
\textbf{Answer Prediction Module:}
A linear layer is applied on the contextual representation $H_L$ to calculate the probability distribution of start and end positions of candidate answer:
\begin{align}
    p^{start} = \text{Softmax}(H_LW^s), \\
    p^{end} = \text{Softmax}(H_LW^e),
\end{align}
where $W^s,W^e\in\mathbb{R}^h$ are learnable parameters.

\noindent
\textbf{Objective Function:} 
The loss function of the backbone MRC is the cross-entropy:
\begin{equation}
\label{loss-rc}
  \min_{\theta_{T,AP}} \mathcal{L}_{span} = -\log p_{y^s}^{start} - \log p_{y^e}^{end},
\end{equation}
where ${y^s}$ and ${y^e}$ are the ground-truth start and end indices of the corresponding sample, respectively.

\subsubsection{Memory Initialization}
For the first domain, to preserve the data distribution of the current domain as much as possible, we randomly select $\mathcal{M}$ training samples from $D_1$ as the initial memory $M_1$.

\subsection{Learning for New Domain}
When the $t$-th domain data is coming, we can access the current data $D_t$ and the memory $M_{t-1}$.
If we just finetune the backbone model, the model is hard to have good transfer ability and may overfit on the few memorized samples.
Hence the MA-MRC utilizes the adversarial learning strategy to fully make use of the domain adaptation relationship between the previous and the current domain.
After the $t$-th training step, we dynamically update memory $M$ with an uncertainty-aware strategy to store the training samples that the model most needs to replay.

\subsubsection{Memory-based Adversarial Training}
To fully use the domain adaption relationship, we adversarially learn the domain-invariant and transfer representations of the memory and current domain data.
Inspired by domain adaptation theory, MA-MRC introduces a domain discriminator and build a two-player min-max game.
The first player, a domain discriminator $D$, distinguishes memory data from the current new domain data. 
Here $D$ is a basic binary discriminator that has a three linear layer followed by a sigmoid activation function.
The second player, the Transformer Encoder $T$, aims to learn features that confuse $D$.
We utilize empirical Maximum Mean Discrepancy as distance measure to reduce the difference between marginal representation distributions and make them similar:
\begin{align}
  {d} = MMD(T(M_{t-1}), T(D_t)).
\end{align}
Finally, this learning procedure can be described by the following $minimax$ game:
\begin{align}
   \min_{\theta_T}& \max_{\theta_D} \mathcal{L}_{adv} = -\frac{1}{\mathcal{M}}\sum\nolimits_{i=1}^{\mathcal{M}}\log D(T(P_i, Q_i)) \nonumber \\ 
    -\frac{1}{|D_t|}&\sum\nolimits_{j=1}^{|D_t|}\log (1-D(T(P_j,Q_j))) + d,
\end{align}
where $\mathcal{M}$ and $|D_t|$ are the number of training samples from memory and the current domain.

\subsubsection{Knowledge Distillation}
It is obvious that a good continual model can learn a new domain data well while approximating the behavior observed in the
old ones.
Then, we leverage knowledge distillation constraints to encourage MA-MRC to mimic its original responses for past samples. 
Specifically, we seek to minimize the distance between the corresponding pre-softmax responses to preserve the knowledge about previous memory data:
\begin{align}
  \mathcal{L}_{KL} = KL(logits_{\theta_{t-1}}^{start} || logits_{\theta_{t}}^{start}) \nonumber \\
  + KL(logits_{\theta_{t-1}}^{end} || logits_{\theta_{t}}^{end}),
\end{align}
where $logits$ are the probability distribution before softmax and $KL$ is the KL divergence.

\subsubsection{Uncertainty-aware Memory Updating}
\label{uncertainty}
Unlike other approaches that store fixed examples for each domain, we use a fixed memory for all domains to avoid memory growing.
Therefore, when a domain arises, it is necessary to remove some old samples in memory to store new data.
We design a dynamic sampling strategy that focuses on samples what the model most needs to replay.
Based on existing trained MRC model $\theta_{t}$, we first compute the uncertainty for each sample in memory by a unsupervised \textit{Entropy-based} stragety:
\begin{align}
  u^{entropy} &= \log p_{y^s}^{start} + \log p_{y^e}^{end}.
\end{align}
Then we calculate the gap between the above uncertainty and 1) the respective previous \textit{best} uncertainty 2) the \textit{average} uncertainty of all memory data, and normalize these metric differentials to create a probability distribution.
For ease of exposition, we define these two normalization calculations as $norm^1$ and $norm^2$.
Note that we will sample $\frac{\mathcal{M}}{t}$ data for each previously seen domain in memory with the above distribution separately. 
Finally, we get $\frac{\mathcal{M}(t-1)}{t}$ data from memory and ramdonly sample $\frac{\mathcal{M}}{t}$ data from current domain.
In this way, the more forgotten a memory sample is, the more it will be retained.

\subsubsection{Objective Function}
When incrementally learning for a new domain, the span loss in answer prediction module $\mathcal{L}_{span}^{'}$ considers all current domain data $D_t$ and the memory data $M$, while the KL loss only takes memory data $M$ into account.
Finally, the overall loss function at the $t$-th step is formulated as:
\begin{align}
   \min_{\theta_{T,AP}} \max_{\theta_{D}} \mathcal{L}_{\text{final}} &= \mathcal{L}_{span}^{'} + \mathcal{L}_{adv} + \mathcal{L}_{KL}.
\end{align}

\begin{table}[t]
\centering
\small
\begin{tabular}{lccccc}
\hline
Domains & $\#$train & $\#$test & $|p|$ & $|q|$ & $|a|$ \\
\hline
\multicolumn{6}{c}{CDA-Q setting} \\
\hline
what & 37593 & 4749 & 118 & 9.9 & 3.5 \\
which & 4159 & 454 & 123.7 & 10.3 & 2.7 \\
where & 3291 & 433 & 120.4 & 8.2 & 3.1 \\
when & 5459 & 696 & 123.3 & 8.6 & 2.3 \\
how & 8124 & 1090 & 120.6 & 9.9 & 3 \\
why & 1201 & 151 & 123.9 & 9.6 & 8 \\
other & 19622 & 1938 & 118 & 11.6 & 2.7 \\
who & 8150 & 1059 & 126.1 & 9 & 2.8 \\
\hline
\multicolumn{6}{c}{CDA-C setting} \\
\hline
SQuAD & 10000 & 10570 & 119.8 & 10.1 & 3.2 \\
NaturalQA & 10000 & 12836 & 152.4 & 9.2 & 4.3 \\
HotpotQA & 10000 & 5901 & 154.1 & 19.5 & 2.3 \\
NewsQA & 10000 & 4212 & 495.1 & 6.6 & 4 \\
TriviaQA & 10000 & 7785 & 674.2 & 13.2 & 1.6 \\
\hline
\end{tabular}
\caption{Dataset statistic of CDA-Q and CDA-C.
  }
\label{data}
\end{table}


\begin{table*}[t]
\centering
\small
\begin{tabular}{lccccccc}
\hline
Methods & SQuAD & NaturalQA & HotpotQA & NewsQA & TriviaQA & $F1_{avg}$& $F1_{all}$ \\
\hline
UpperBound & 78.24 & 68.66 & 68.03 & 58.63 & 63.47 & 67.41 & 69.02 \\
LowerBound & 62.99 & 56.37 & 58.93 & 42.35 & 64.52 & 57.03 & 58.53 \\
\hline
EWC & 59.28 & 53.38 & 57.62 & 41.56 & 65.81 & 55.53 & 56.63 \\
OnlineEWC & 67.64 & 58.02 & 59.64 & 46.68 & 65.30 & 59.46 & 60.93 \\
AGEM* & 63.22 & 56.85 & 57.94 & 46.62 & 65.37 & 58.00 & 59.20 \\
DER* & 41.53 & 64.98 & 63.24 & 56.48 & 64.56 & 58.16 & 57.78 \\
DERPlus++* & 46.96 & 64.19 & 63.53 & 56.62 & 63.94 & 59.05 & 58.87 \\
MA-MRC($norm^1$)* & 68.13 & 61.88 & 61.29 & 52.20 & 66.42 & \textbf{61.98} & \textbf{63.26} \\
MA-MRC($norm^2$)* & 70.11 & 60.83 & 61.96 & 52.18 & 65.64 & \textbf{62.14} & \textbf{63.39} \\
\hline
\end{tabular}
\caption{The overall results under CDA-C setting. ``*'' indicates that the $\mathcal{M}$ memory data is used.}
\label{overall_C}
\end{table*}

\begin{table*}[t]
\centering
\small
\begin{tabular}{lcccccccccc}
\hline
Methods & what & which & where & when & how & why & other & who & $F1_{avg}$& $F1_{all}$ \\
\hline
UpperBound & 83.86 & 85.16 & 81.03 & 92.36 & 83.36 & 65.30 & 84.22 & 88.82 & 83.01 & 84.60 \\
LowerBound & 68.92 & 76.65 & 61.87 & 67.06 & 74.55 & 30.25 & 75.43 & 87.17 & 67.74 & 71.89 \\
\hline
EWC & 68.72 & 77.86 & 58.95 & 63.92 & 72.21 & 41.51 & 72.46 & 86.82 & 67.81 & 70.87 \\
OnlineEWC & 71.23 & 76.53 & 66.81 & 63.13 & 69.89 & 38.74 & 75.10 & 87.60 & 68.63 & 72.49 \\
AGEM* & 71.08 & 77.37 & 72.37 & 77.87 & 70.74 & 57.19 & 76.03 & 88.98 & 73.95 & 74.32 \\
DER* & 58.29 & 82.48 & 79.13 & 90.98 & 84.03 & 66.53 & 84.17 & 89.06 & 79.33 & 72.93 \\
DERPlus++* & 62.26 & 79.75 & 79.17 & 90.07 & 82.61 & 63.63 & 84.37 & 88.59 & 78.80 & 74.34 \\
MA-MRC($norm^1$)* & 78.87 & 80.59 & 77.87 & 90.19 & 81.81 & 67.36 & 83.13 & 90.09 & \textbf{81.24} & \textbf{81.69} \\
MA-MRC($norm^2$)* & 78.63 & 81.80 & 77.24 & 89.35 & 79.94 & 63.83 & 83.23 & 89.29 & \textbf{80.41} & \textbf{81.25} \\
\hline
\end{tabular}
\caption{The overall results under CDA-Q setting. ``*'' indicates that the $\mathcal{M}$ memory data is used.}
\label{overall_Q}
\end{table*}

\section{Experiment}
\subsection{Continual MRC Datasets}
Inspired by \citet{DBLP:conf/cikm/SuGZFLC20}, this paper deals with two continual MRC under domain adaptation: CDA-C and CDA-Q. 
For CDA-C setting, we regard MRC datasets with different passage corpora (e.g., Wikipedia, News, and Web snippets) as different domains
and choose five datasets: SQuAD 1.1 \cite{RajpurkarZLL16}, HotpotQA \cite{Yang0ZBCSM18},  Natural Questions \cite{TrischlerWYHSBS17}, NewsQA \cite{TrischlerWYHSBS17}, and TriviaQA \cite{DBLP:conf/acl/JoshiCWZ17}.
Due to computational limits, we use the curated version provided by \citet{DBLP:conf/acl-mrqa/FischTJSCC19}.
For each dataset, we randomly sample 10,000  $<$question, context, answer$>$ triples from the original training datasets for continual training, and the original dev sets for testing.
For CDA-Q setting, we make use of the original SQuAD 1.1 \cite{RajpurkarZLL16} and split it into eight domains according to the question type such as what, why, and how.
The detailed statistics of these datasets are shown in Table \ref{data}.

\begin{table*}[t]
\centering
\small
\begin{tabular}{lccccccc}
\hline
Methods & SQuAD & NaturalQA & HotpotQA & NewsQA & TriviaQA & $F1_{avg}$& $F1_{all}$ \\
\hline
Full Model & 68.13 & 61.88 & 61.29 & 52.20 & 66.42 & 61.98 & 63.26 \\
{ }w/o Adv & 63.79 & 57.40 & 57.76 & 48.36 & 65.27 & 58.51 & 59.65 \\
{ }w/o KL & 62.65 & 58.90 & 60.38 & 51.48 & 66.41 & 59.96 & 60.73 \\
{ }w/o Adv+KL & 53.10 & 50.73 & 55.14 & 47.06 & 64.45 & 54.10 & 54.18 \\
\hline
\end{tabular}
\caption{The ablation study results under CDA-C setting.}
\label{ablation_C}
\end{table*}

\begin{table*}[t]
\centering
\small
\begin{tabular}{lcccccccccc}
\hline
Methods & what & which & where & when & how & why & other & who & $F1_{avg}$& $F1_{all}$ \\
\hline
Full Model & 78.87 & 80.59 & 77.87 & 90.19 & 81.81 & 67.36 & 83.13 & 90.09 & 81.24 & 81.69 \\
{ }w/o Adv & 73.33 & 78.86 & 73.45 & 88.12 & 77.83 & 62.29 & 76.87 & 89.74 & 77.56 & 77.15 \\
{ }w/o KL & 77.67 & 82.86 & 78.73 & 89.03 & 81.83 & 66.40 & 83.21 & 90.04 & 81.22 & 81.21 \\
{ }w/o Adv+KL & 73.15 & 76.90 & 65.96 & 84.99 & 74.62 & 55.76 & 76.94 & 88.07 & 74.55 & 75.89 \\
\hline
\end{tabular}
\caption{The ablation study results under CDA-Q setting.}
\label{ablation_Q}
\end{table*}

\subsection{Methods Compared}
\textbf{$\bullet$ Bounds.} 
We design a standard BERT-MRC (the same as the backbone MRC model in Sec \ref{backbone}) as the basic model and then define two bounds on it:
1) \textbf{Lower Bound} continually fine-tunes the BERT-MRC model for each new domain without memorizing any historical examples.
2) \textbf{Upper Bound} remembers all examples in history and continually re-train the BERT-MRC model with all data.

\textbf{$\bullet$ EWC} \cite{kirkpatrick2017overcoming}
  restricts the change of model parameters for previous domains via elastic weight consolidation and a special $L_2$ regularization.
  Hence it can slow down the learning of parameters important for all previous domains.

 \textbf{$\bullet$ Online EWC} \cite{DBLP:conf/icml/Schwarz0LGTPH18},
  the extension of EWC, which only consider the restriction for the latest model parameters.
 
 \textbf{$\bullet$ DER} \cite{DBLP:conf/nips/BuzzegaBPAC20},
  memory-based approaches, leverages knowledge dsistillation for retaining past experience.

 \textbf{$\bullet$ DER++} \cite{DBLP:conf/nips/BuzzegaBPAC20}, the extension of DER, uses an additional term on memory.

 \textbf{$\bullet$ AGEM} \cite{DBLP:conf/iclr/ChaudhryRRE19},
  memory-based approaches, uses a constraint that enables the projected gradient to decerease the average loss on previous seen domains.

\subsection{Evaluation Metrics}
Exact Match (EM) and word-level F1 score are used to evaluate the performance of MRC model in a single domain data.
As for the continual domain adaptation setting, two common evaluation settings in continual learning theory are adopted: the average and the whole performance:
\begin{align}
  F1_{avg} &= \frac{1}{T}\sum_{i=1}^TF1(D_{test}^i), \\
  F1_{all} &= F1(D_{test}^{1:T}).
\end{align}
The former is the average F1 score on test sets of all seen domain, and the latter is the whole F1 score on the test sets.

\subsection{Implementation Details}
We initialize the transformer encoder layer with the pre-trained $\text{BERT}_\text{BASE}$ model officially released by Google\footnote{https://github.com/google-research/bert}.
The maximum sequence length is 384, and the batch size is 20.
We set memory size $\mathcal{M}$ 400 default that means the memory stores up to 400 training samples for previous seen domain.
When incrementally learn the new incoming domain data at $t$-th step, we first reinitialize the parameters of the domain discriminator $\theta_D$.
Then use Adam optimizer \cite{KingmaB14} with a learning rate of 3e-5 and training MA-MRC model for 3 epochs.



\begin{figure}[h!]
\centering
  \subfigure[Under CDA-C setting.]{
  \includegraphics[width=0.9 \linewidth]{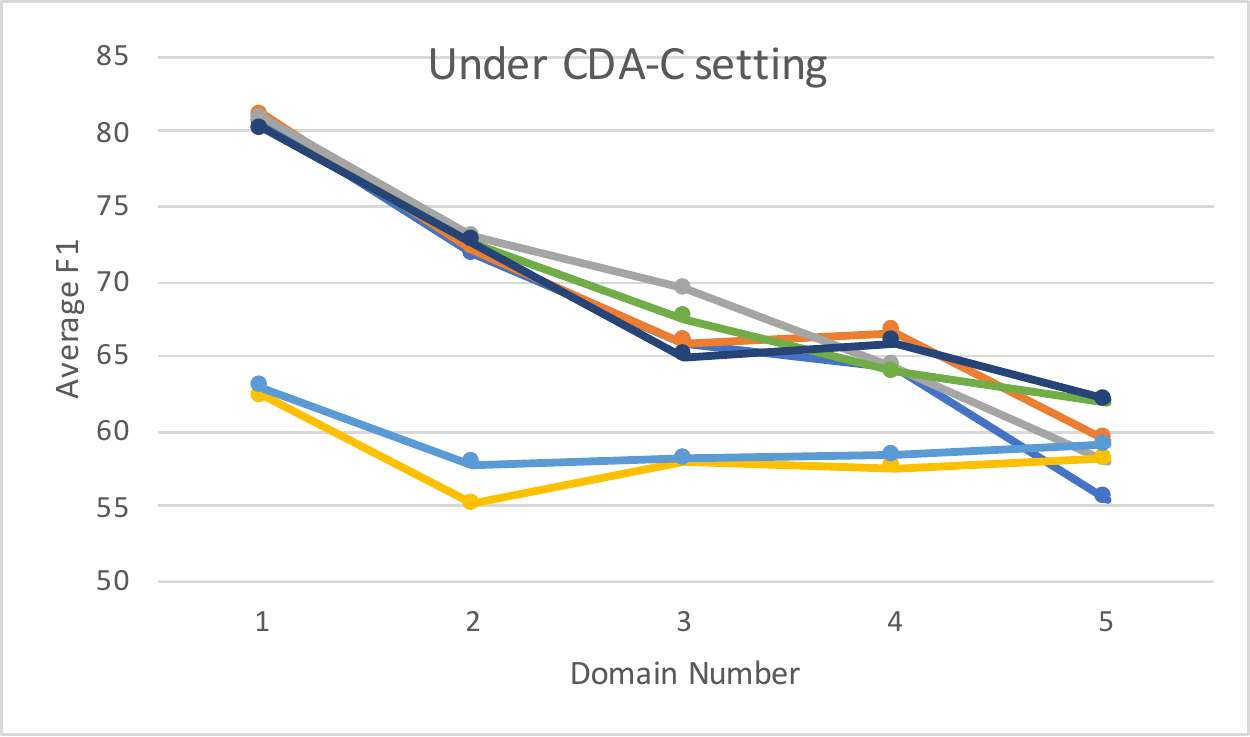}
  \label{tsne1}
  }
  \subfigure[Under CDA-Q setting.]{
  \includegraphics[width=0.9 \linewidth]{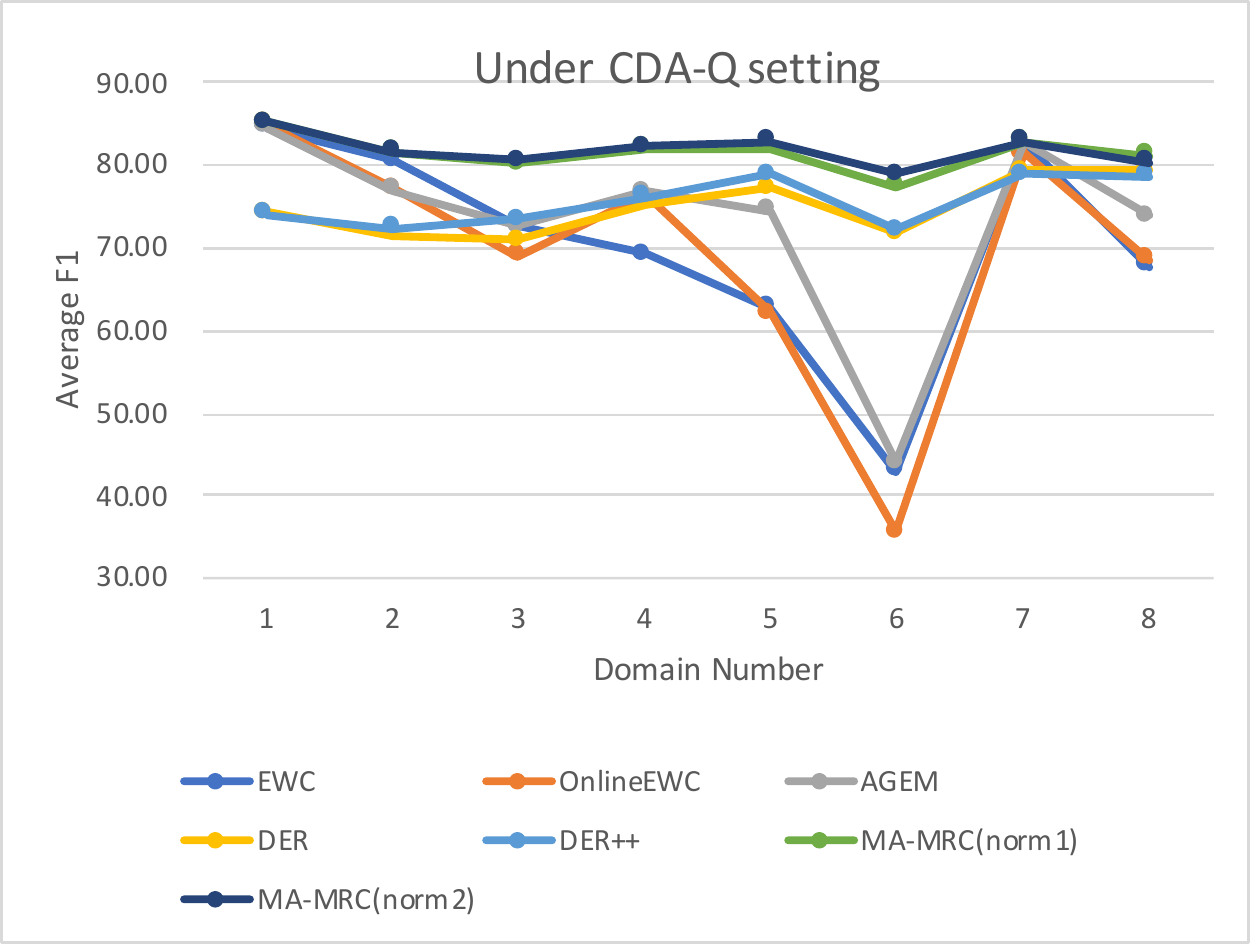}
  \label{tsne2}
  }
  \caption{The average F1 changes with increasing domains through the continual learning process.}
  \label{plot_all_f1}
\end{figure}
\subsection{Results and Discussion}
\subsubsection{Main Results}

For the task of continual MRC, the overall results under CDA-C and CDA-Q setting and are shown in Table \ref{overall_C} and Table \ref{overall_Q}.
Taking $F1_{avg}$ and $F1_{all}$ as two overall performance evaluations, we have the following findings:
(1) The proposed MA-MRC methods outperform other baselines by a significant margin on both two continual settings.
(2) The performance trend of different methods are not consistent across two continual settings. 
For example, OnlineEWC performs well under CDA-C but performs poorly CDA-Q, and all memory-based methods outperform consolidation-based methods under CDA-Q but not CDA-C.
The reason is that the characteristics of the domains and the continual learning difficulty are different.
Concretely, it is obvious that the domain data in CDA-C is more different and more difficult than that in CDA-Q.
(3) There is a big gap between all the models and the upper bound.
We further demonstrate the evaluation results of the proposed MA-MRC and UpperBound methods on each domain at each continual learning step $t$ in Table \ref{all}.
The results indicate that MA-MRC indeed forgets knowledge learned on previously seen domain data, and there remain lots of things to be explored for continual MRC.

\begin{table}[t]
\small
\begin{tabular}{llllll}
\hline
 $Data$ & $D_1$ & $D_2$ & $D_3$ & $D_4$ & $D_5$ \\
\hline
\multicolumn{3}{l}{MA-MRC:} \\
step 1: & 80.13 & - & - & - & - \\
step 2: & 74.92 & 70.32 & - & - & - \\
step 3: & 64.94 & 60.81 & 69.28 & - & - \\
step 4: & 74.13 & 63.94 & 65.14 & 60.36 & - \\
step 5: & 70.11 & 60.83 & 61.96 & 52.18 & 65.64 \\
\hline
\multicolumn{3}{l}{UpperBound:} \\
step 1: & 81.08 & - & - & - & - \\
step 2: & 79.51 & 69.94 & - & - & - \\
step 3: & 78.42 & 68.96 & 70.06 & - & - \\
step 4: & 76.91 & 67.63 & 67.52 & 59.36 & - \\
step 5: & 78.24 & 68.66 & 68.03 & 58.63 & 63.47\\
\hline
\multicolumn{3}{l}{MA-MRC w/o Adv:} \\
step 1: & 80.752 & - & - & - & - \\
step 2: & 76.495 & 71.395 & - & - & - \\
step 3: & 72.348 & 64.178 & 70.568 & - & - \\
step 4: & 65.005 & 56.839 & 60.652 & 60.685 & - \\
step 5: & 63.788 & 57.397 & 57.761 & 48.355 & 65.269 \\
\hline
\end{tabular}
\caption{Catastrophic forgetting phenomenon of proposed MA-MRC(w/o Adv) and UpperBound mehtods under CDA-C setting.}
\label{all}
\end{table}

\begin{table*}[t]
\centering
\small
\begin{tabular}{lccccccc}
\hline
Sampling & SQuAD & NaturalQA & HotpotQA & NewsQA & TriviaQA & $F1_{avg}$& $F1_{all}$ \\
$u^{entropy}, norm^1$ & 68.13 & 61.88 & 61.29 & 52.20 & 66.42 & 61.98 & 63.26 \\
$u^{entropy}, norm^2$ & 70.11 & 60.83 & 61.96 & 52.18 & 65.64 & 62.14 & 63.39 \\
$u^{prob}, norm^1$ & 68.71 & 60.81 & 62.05 & 52.79 & 66.27 & 62.13 & 63.22 \\
$u^{prob}, norm^2$ & 63.78 & 61.74 & 59.91 & 52.72 & 65.92 & 60.81 & 61.87 \\
Random & 65.68 & 58.61 & 59.84 & 52.47 & 65.99 & 60.52 & 61.36 \\
\hline
\end{tabular}
\caption{The results of different sampling straties under CDA-C setting.}
\label{sample_C}
\end{table*}

\begin{table}[t]
\centering
\small
\begin{tabular}{llll}
\hline
$\mathcal{M}$ & methods & $F1_{avg}$ & $F1_{all}$ \\
\hline
 & MA-MRC($norm^1$) & 55.26 &  55.52  \\
\multirow{-2}{*}{200} & MA-MRC($norm^2$) & 57.72 & 58.35  \\
\hline
 & MA-MRC($norm^1$) & 60.00  & 60.79  \\
\multirow{-2}{*}{300} & MA-MRC($norm^2$) & 59.73 & 60.59  \\
\hline
 & MA-MRC($norm^1$) & 61.98  & 63.26  \\
\multirow{-2}{*}{400} & MA-MRC($norm^2$) & 62.14 & 63.39  \\
\hline
\end{tabular}
\caption{Results with different $\mathcal{M}$ under CDA-C.}
\label{results_m}
\end{table}

\begin{table}[t]
\centering
\small
\begin{tabular}{llll}
\hline
 &  & $F1_{avg}$ & $F1_{all}$ \\
\hline
 & DER++ &  75.85 &  67.94 \\
 & AGEM & 40.34 &  42.07 \\
 & MA-MRC($norm^1$) & 79.05  & 78.09 \\
\multirow{-4}{*}{order 1} & MA-MRC($norm^2$) &  79.98 &  79.11  \\
\hline
 & DER++ &  77.06 &   83.22 \\
 & AGEM & 81.73 &  83.91 \\
 & MA-MRC($norm^1$) & 82.37  & 84.41 \\
\multirow{-4}{*}{order 2} & MA-MRC($norm^2$) & 82.32  & 84.91  \\
\hline
 & AGEM & 73.83 &  73.46  \\
 & MA-MRC($norm^1$) & 79.64  & 81.02  \\
\multirow{-3}{*}{order 3} & MA-MRC($norm^2$) & 78.94 & 80.47   \\
\hline
\end{tabular}
\caption{The results of different domain order under CDA-Q setting.}
\label{order}
\end{table}

\begin{table}[t]
\centering
\small
\begin{tabular}{llcc}
\hline
  & & \multicolumn{2}{l}{Training Time / 1 epoch} \\
  & \multirow{-2}{*}{Para} & CDA-C & CDA-Q\\
\hline
  AGEM & $\theta_{T,AV}$ & 77 m & 49 m \\
  DER & $\theta_{T,AV}$ & 57 m & 38 m \\
  DER++ & $\theta_{T,AV}$ & 56 m & 38 m  \\
  MA-MRC($norm^1$) & $\theta_{T,AV,D}$ & 61 m & 40 m  \\
  MA-MRC($norm^2$) & $\theta_{T,AV,D}$ & 60 m & 39 m  \\
\hline
\end{tabular}
\caption{Parameters and speed comparison.}
\label{speed}
\end{table}

Besides, we plot the average F1 performance of models during the whole continual learning process in Figure \ref{plot_all_f1} to investigate how performance changes.
We observe that the performance of all models decreases in some degree with increasing numbers of domains under both CDA-C and CDA-Q settings.
However, the proposed MA-MRC methods outperform other baselines and achieve better performance on the whole domain data.

\subsubsection{Ablation Study}
To better understand our proposed model, we conduct ablation studies to see the effectiveness of each model component.
The results in Table \ref{ablation_C} and Table \ref{ablation_Q} demonstrate that both knowledge distillation and adversarial training contribute to avoiding largely forgetting.
For incremental domain learning, knowledge distillation is a naive way to enforce the model to remember its original responses for previous domains.
What is more, the adversarial domain adaptation, can make MRC model learn domain-invariant and transfer representations better.
Table \ref{all} shows the evaluation results of w/o Adv at each continual step.
The results indicate that the w/o Adv method has a more harmful forgetfulness on the very previous seen domain than MA-MRC model.
The results prove that the adversarial learning indeed helps for remembering previous knowledge.

\subsubsection{Effect of Memory}
\textbf{Memory Size $\mathcal{M}$.} \quad 
Table \ref{results_m} shows the performance with three different memory size $\mathcal{M}$: 200, 300, and 400.
In low memory scenario, i.e., $\mathcal{M}=200$, the proposed method performs poorly.
The reason is that the adversarial domain adaption has difficulty transferring well with a too small memory.
Therefore, as the number of memory samples increases, it will be more conducive to transfer to a new domain, and the overall performance will be better.
We believe that an appropriate memory size could lead to better performance.

\noindent
\textbf{Uncertainty-aware Sampling.} \quad 
We replace the uncertainty-aware memory updating strategy with another two strategies.
First, we use another uncertainty measurement that takes heuristic max softmax probability of spans as the uncertainty: $u^{prob} = \max_{i,j} (p_i^{start} + p_j^{end})$.
The second strategy is random sampling.
The experimental results in Table \ref{sample_C} indicate that the uncertainty-aware sampling (both \textit{Entropy-based} and \textit{Probability-based} uncertainty) is better than random sampling.

\subsubsection{Effect of Domain Order}
Table \ref{order} shows the results of different domain orders.
Order 1 is a descending order based on the number of training samples in each domain and order 2 is an ascending order, and order 3 is a random order.
The performance of AGEM and DER++ degrades severely in order 1.
However, the proposed MA-MRC methods is superior to baselines and are stable and robust under different orders.


\subsubsection{Efficiency Analysis}
We compare the parameters and training speed of methods with the same size of the memory in Table \ref{speed}.
The MA-MRC have more parameters for domain discriminator.
Nevertheless, considering the larger number of parameters of the Transformer, we conclude all methods have almost the same number of parameters.
As for training time per epoch, MA-MRC is slower than DER++ (4\%/5\% under CDA-C/CDA-Q).
It can be accepted because of the significant improvement of MA-MRC.


\section{Conclusion}
In this paper, an incremental learning MRC model with uncertainty-aware fixed memory and adversarial domain adaptation, MA-MRC, is proposed for continual MRC and alleviating catastrophically forgetting.
Inspired by the human learning process, There are two main ideas of MA-MRC: 
a memory that stores a small number of samples in previous seen domain data and always focuses on what the model most needs to replay;
and adversarial learning the domain adaptation in a two-player game to learn better transfer representations between previous and current domain data.
Experimental results show that the proposed MA-MRC can achieve a good continuous learning performance without catastrophically forgetting under CDA-C and CDA-Q settings.

In the future, we would like to explore a more effective sampling strategy, domain adaptation strategy, and balance training strategy for multiple objectives to enhance the continual MRC model.

\section*{Acknowledgment}
This paper is funded by National Natural Science Foundation of China (Grant No. 62173195).
This work is partly supported by seed fund of Tsinghua University (Department of Computer Science and Technology) -Siemens Ltd., China Joint Research Center for Industrial Intelligence and Internet of Things.

\bibliography{anthology,custom}
\bibliographystyle{acl_natbib}




\end{document}